\documentclass[10pt,twocolumn,letterpaper]{article}

\usepackage[pagenumbers]{cvpr} % To force page numbers, e.g. for an arXiv version

\usepackage{graphicx}
\usepackage{amsmath}
\usepackage{amssymb}
\usepackage{booktabs}
\usepackage{graphicx}
\graphicspath{ {./Images/} }
\usepackage[export]{adjustbox}
\usepackage{svg}
\usepackage[accsupp]{axessibility}

\usepackage{fancyhdr}
\usepackage[pagebackref,breaklinks,colorlinks]{hyperref}

\usepackage[capitalize]{cleveref}
\crefname{section}{Sec.}{Secs.}
\Crefname{section}{Section}{Sections}
\Crefname{table}{Table}{Tables}
\crefname{table}{Tab.}{Tabs.}

\def\cvprPaperID{12314} % *** Enter the CVPR Paper ID here
\def\confName{CVPR}
\def\confYear{2024}

\begin{document}

\title{Insights from the Use of Previously Unseen Neural Architecture Search Datasets}

\author{Rob Geada\thanks{These authors performed equal contribution.} \; David Towers\textsuperscript{*, 1}  Matthew Forshaw\textsuperscript{1,3}  Amir Atapour-Abarghouei\textsuperscript{2}  A. Stephen McGough\textsuperscript{1}\\
    \and
    \textsuperscript{1}Newcastle University, UK --- \textsuperscript{2}Durham University, UK --- \textsuperscript{3}The Alan Turing Institute, UK\\
    {\tt\small rob@geada.net, amir.atapour-abarghouei@durham.ac.uk}\\
    {\tt\small \{d.towers2, matthew.forshaw, stephen.mcgough\}@ncl.ac.uk} \\
}
\maketitle
\thispagestyle{fancy}

%%%%%%%%% ABSTRACT
\begin{abstract}
    The boundless possibility of neural networks which can be used to solve a problem -- each with different performance -- leads to a situation where a Deep Learning expert is required to identify the best neural network. This goes against the hope of removing the need for experts. Neural Architecture Search (NAS) offers a solution to this by automatically identifying the best architecture. However, to date, NAS work has focused on a small set of datasets which we argue are not representative of real-world problems.
    We introduce eight new datasets created for a series of NAS Challenges: AddNIST, Language, MultNIST, CIFARTile, Gutenberg, Isabella, GeoClassing, and Chesseract. These datasets and challenges are developed to direct attention to issues in NAS development and to encourage authors to consider how their models will perform on datasets unknown to them at development time. We present experimentation using standard Deep Learning methods as well as the best results from challenge participants.\vspace{-0.5cm}

   %We introduce Neural Architecture Search Datasets from the Unseen-data Challenge (NAS-DUCs). This collection includes eight datasets that have been developed for the NAS Unseen-Data competition held at CVPR 2021, 2022, and 2023; AddNIST, Language, MultNIST, CIFARTile, Gutenberg, Isabella, GeoClassing, and Chesseract. These datasets, and the competition they were a part of, were developed to direct attention to an issue in NAS development, the overuse of standard benchmark datasets, and to encourage authors to consider how their models will perform on datasets unknown to them. We present experimentation using standard deep learning methods and the best results from methods submitted to the competition.
\end{abstract}

%%%%%%%%% BODY TEXT
\section{Introduction}
\label{sec:intro}
One of the main appeals of Deep Learning (DL) was its ability to democratise Machine Learning. No longer would you require a domain expert to develop an optimal solution to a given problem since DL models are capable of learning the required patterns and features from the data themselves and would not need the domain expert to identify and extract the required features. Unfortunately, rather than removing the need for an expert, the new paradigm has just shifted where the expert is required. These experts now spend their time identifying the most optimal neural network architecture for a given problem. Those who are not that proficient would select between pre-existing off-the-shelf networks such as ResNet~\cite{he_deep_2016}, VGG~\cite{k_simonyan_very_2014} or Inception~\cite{christian_szegedy_going_2015}. However, to get the best results for a particular problem, one should search across all possible solutions, not just those which may have shown good results in other problem domains, as no network is optimal for all problems. This idea of the `no free lunch'~\cite{david_h_wolpert_no_1997} is backed up by our findings here, with some networks giving far poorer performance than what would be expected.

A DL problem can be seen as the combination of a particular task -- such as classification -- one wishes to apply in a particular data domain. The data domain is normally realised as a dataset for training the DL network.

Neural Architecture Search (NAS) is a fast-developing field of DL -- which can be seen from the increase of published papers in the domain over the last few years \cite{colin_white_neural_2023}. The aim of NAS is to remove the need for an expert's time and knowledge to generate state-of-the-art competitive Neural Networks from a particular dataset \cite{zoph_neural_2016}. NAS methods search through millions (if not billions) of candidate architectures from a pre-defined search space to find optimum networks. These search spaces comprise not only different network topologies but also the types of nodes which make up these networks -- such as fully connected layers, convolutions and pooling. Although NAS can be applied to any data modality, the majority of work to date has focused on image-based datasets, with modern NAS methods achieving this feat in a few GPU hours\cite{mellor_neural_2021, peng_ye_beta-darts_2022}. In addition to a pre-defined search space, a NAS approach will have a search strategy. In terms of the task, the majority of work in NAS has focused on classification problems.

NAS techniques are often evaluated by their ability to find networks that perform well on benchmark datasets, such as CIFAR-10 \cite{krizhevsky_learning_2009} and ImageNet \cite{jia_deng_imagenet_2009}. These datasets, combined with standard search spaces such as NAS-Bench \cite{dong_nats-bench_2021}, fill an important role, allowing a direct comparison between NAS methods. While there is a clear benefit in using common benchmark datasets, we argue this is against the ethos of NAS -- removing the need for domain experts, required to develop optimal architectures for a given problem.

Developing against common datasets can be seen as over-engineering a process to reach a solution we already know. For example, for datasets such as CIFAR-10, extensive experimentation with various neural network architectures has led to extremely high levels of accuracy \cite{dosovitskiy_image_2020, maxime_oquab_dinov2_2023, andrea_gesmundo_evolutionary_2022}, rendering this dataset easily solvable. In essence, since we already know what achieves high performance, NAS strategies may be optimised to reach these solutions, which may not generalise to other datasets. Prior work \cite{yash_mehta_nas-bench-suite_2022} has found that just because NAS methods perform well on benchmarks such as NAS-Bench-101\cite{ying_nas-bench-101_2019}, this does not mean they will match other benchmarks.

%While the benefit of these common benchmark datasets and other similar datasets cannot be understated, the utility of NAS is to eliminate the need for an expert's time and knowledge during model design, which means it needs to be generalisable to any problem. 

If NAS is to truly replace architecture design experts, it must be able to find optimal networks for datasets that it has never seen before -- the unseen data challenge. There is a dichotomy here in that proposing a new `unseen' dataset for NAS may lead to that dataset being adopted as one of the core datasets for NAS. We, therefore, see our work as having two values. In the first case, we are diluting the effect of overfitting to the core datasets -- having more core datasets is less likely to create a NAS approach, which can overfit to all of them. For the second case, we see this as a `call to arms' to encourage the development and use of more datasets for NAS, which we will continue to do ourselves.

%If a model architecture created via NAS is to be deployed in the real world, it needs to be able to generalise to unseen data. However, most NAS algorithms list results exclusively on common benchmark datasets and often use standard search spaces, which are known to contain good architectures for these datasets; this is because work on manually designed models for those exact standard benchmark datasets have provided a set of best-practices, defining what operations and design patterns work best. However, will design choices that produce good models for datasets like CIFAR-10, for instance, necessarily produce good models for other datasets? Results on the datasets discussed in this paper by standard CNN models show that while one method may be better overall, it won't necessarily be the best every time.
%We believe that the overuse of these benchmark datasets has influenced the direction of NAS. NAS methods are good at generating architectures for these well-known datasets but this may not necessarily transfer to other novel datasets.

In this work, we present eight new datasets created and used for unseen dataset NAS challenges. We define how these datasets were constructed and illustrate benchmark performance metrics that have been achieved, with the intention of encouraging others in the NAS community to develop approaches that will work on more general datasets.

\section{Motivation}
\label{sec:mot}

% To understand how well NAS methods might generalise to novel datasets, We ran the ``Unseen-data Challenge'' at the Second, Third, and Fourth workshops on Neural Architecture Search, which we helped organise at CVPR 2021, 2022, and 2023, respectively. In these challenges, participants were tasked to create NAS methods to perform well on unknown datasets and tasks. 

To understand how well NAS methods might generalise to novel datasets and drive the direction of NAS research towards more impactful real-world applications, we organised a challenge tasking participants with creating NAS methods that perform well on unknown datasets.
Participants were provided with a small number of ``development" datasets to develop and test their NAS algorithm(s). After development, the algorithms were submitted to the competition servers, where the algorithms would be run over novel ``evaluation" datasets. The datasets were created bespoke for the challenge and withheld from the participants throughout its duration, such that no participant could gain an advantage by knowing or guessing the evaluation datasets. As such, the participants are encouraged to create NAS approaches that generalise well to multiple problems rather than ones specifically targeted at one dataset.

These unseen datasets explore one of two concepts: \vspace{-0.2cm}
\begin{itemize}
    \item{Type-1:} A problem an expert could solve themselves or create a program, which may include basic DL, that can outperform a naive DL approach on the raw datasets. An example of this type of data is the AddNIST dataset \cref{desc:addnist}, one of the datasets outlined later in this paper, which consists of images with three channels, with each channel being an image from the MNIST dataset \cite{yann_lecun_mnist_2005} having the class label as the sum (the numbers are chosen so the sum is less than twenty) of the individual MNIST digits. With prior knowledge of this dataset, a solution that attains the highest accuracy is splitting the colour channels into three separate images, using a model trained on MNIST to identify the individual numbers, and then hardcoding in the necessary arithmetic to get the final answer.\vspace{-0.2cm}
    \item{Type-2:} A problem that would be almost impossible for a human to solve or create a program without task-specific tools. An example is the Language dataset \cref{desc:lang}, which encodes words from ten languages into an image. The models needs to identify which letter frequency is attributed to the correct language. Without prior knowledge of what is encoded within the images and which language is represented by each class label, correctly labelling each image to the correct language would be nearly impossible. The motivation for using a type 2 dataset is to see if bespoke models can perform better than random guesswork over datasets that appear human-impossible at face value. 
\end{itemize}\vspace{-0.2cm}

We believe that both types of tasks are essential for NAS algorithms to make a meaningful impact in real-world applications. First, if NAS remains incapable of solving problems that are effortlessly solvable by humans, it would be premature to assert that NAS can remove the requirement for expert knowledge. The second type relates to how machine learning is often deployed to find solutions to problems humans find complicated. Essentially, NAS methods should be able to perform at a level at least commensurate with other conventional machine learning techniques to effectively supplant manually designed models.

To further obfuscate the true nature of the datasets from the challenge participants, within our datasets, data shapes were deliberately chosen that \textit{appeared} to correlate to normal images or well-known datasets, such as 3$\times$64$\times$64, and data split sizes to align with well-known benchmark datasets, such as 50,000 training images to align with CIFAR. We also asked participants for their guesses as to what each development dataset was: most participants correctly described the Type-1 datasets. In contrast, none of the  Type-2 datasets have been correctly identified. \vspace{-0.2cm}  

\section{Related Work}

We consider related work within the NAS literature and the datasets commonly used in NAS research.

\subsection{Neural Architecture Search}

NAS aims to automate the process of neural network architecture design. Traditional network design requires extensive human expertise and significant time investment. NAS seeks to streamline this process by employing algorithms that efficiently identify architectures tailored to specific tasks and datasets from a vast search space.

Initially, the work by Zoph and Le~ \cite{zoph_neural_2016} applied NAS using reinforcement learning to CIFAR-10, achieving an architecture slightly outperforming manually designed models. NASNet \cite{zoph_learning_2018} addressed the challenge of learning architectures directly on large datasets by transferring a building block designed for a small dataset to a larger one using ImageNet. ENAS \cite{hieu_pham_efficient_2018} focused on computational efficiency by learning to search for an optimal sub-graph within a large graph, thus reducing computational resources. Evolutionary algorithms \cite{esteban_real_large-scale_2017} have emerged, providing comparable results but with a distinct evolutionary overview. Weight-sharing methods \cite{li_random_2019} define over-parameterised super-networks (one-shot models), reducing computational costs, allowing the training of one super-network encompassing many different sub-architectures rather than training numerous networks independently.

Differentiable NAS \cite{hanxiao_liu_darts_2018,geada_bonsai-net_2020} marks another significant advancement, allowing gradient-based optimisation methods to explore the neural architecture search space efficiently. Each of these developments in NAS represents a significant stride towards the more efficient, automated and effective design of neural network architectures; however, the effectiveness of these approaches is often only tested on common benchmark datasets, some of which we will briefly cover in the following sections. The real-world applicability of these approaches needs to be tested on ``unseen" datasets, which better simulate what is needed of NAS in practice.

\subsection{Common NAS datasets}
Although any dataset can be used to evaluate NAS methods, CIFAR-10 \cite{krizhevsky_learning_2009}, and ImageNet \cite{jia_deng_imagenet_2009} are highly popular.

CIFAR-10 \cite{krizhevsky_learning_2009} comprises 60,000 images from ten classes evenly distributed in training and testing sets. Accuracy on CIFAR-10 is often used to evaluate performance, and it is also used in NAS, with the very first NAS algorithm \cite{zoph_neural_2016} highlighting the potential power of NAS by demonstrating its ability to achieve better CIFAR-10 performance than human-designed models. While CIFAR-10 is a good dataset for comparing performance, it is an easy problem, with modern architectures and NAS methods easily achieving accuracies over 90\%. Furthermore, since CIFAR-10 is used as a standard benchmark, new methods are encouraged to focus on this dataset, which may come at a cost of generalisability across other datasets.

ImageNet \cite{jia_deng_imagenet_2009} is another popular dataset for evaluation and benchmarking. It contains 1,000 classes and presents a more challenging problem, with the results often presented in tandem with CIFAR-10 results to demonstrate model performance. ImageNet poses similar issues to CIFAR-10 due to its similar prevalence in benchmarking DL methods. While including ImageNet and CIFAR-10 as datasets for NAS comparison demonstrates that the NAS approach is not overfitting to just one dataset, it does not solve the problem of generalisation. We argue that both more datasets and datasets that the developers are unaware of are needed. %ImageNet is considered a similar problem to CIFAR-10.

\subsection{NAS Benchmark Suites}
To make NAS development more comparable and easier, benchmark suites such as NAS-Bench-101 \cite{ying_nas-bench-101_2019}, NAS-Bench-201 \cite{dong_nas-bench-201_2020}, and NATS-Bench \cite{dong_nats-bench_2021} have been developed. In these benchmarks, the performance achieved for each network in the search space has been pre-computed, thus removing the model training step in NAS approaches.

Benchmark suites allow developers to focus on the development of search strategies and allow for easy comparison between different approaches as they work directly on the same search space. Thus, performance gains over other methods cannot be attributed to different search spaces.

These benchmark suites suffer from the same issues of reliance on CIFAR-10 and ImageNet. The search spaces used by these benchmarking tools have been fully explored on CIFAR-10 and ImageNet. These results are stored so a NAS method can look up an architecture's final result without having to retrain fully. These lookup tables do not exist for other datasets, meaning that NAS methods developed using these tools can quickly generate results for CIFAR-10 and ImageNet but not easily for other datasets.

\section{Dataset Descriptions}

\begin{figure*}
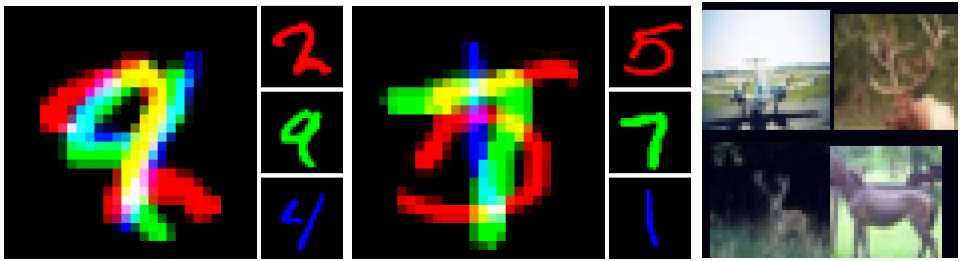
 
    \centering

    \includesvg[width=0.26\textwidth]{Images/AddNIST_single.svg}
    \includesvg[width=0.26\textwidth]{Images/MultNIST_single.svg}
    \includesvg[width=0.2\textwidth]{Images/CIFARTile_single.svg}
    \caption{\textit{Left:} AddNIST image - the sum of the channels adds up to 15 (r = 2, g = 9, b = 4), implying a label of 14. \textit{Middle} MultNIST image - the product of the channels equals 35 (r = 5, g = 7, b = 1), which implies a label of 5 (35 \%10 = 5). \textit{Right} CIFARTile image - two deer, one aeroplane, and one horse, meaning there are three unique labels among the sub-images, which equates to a final label of 2}
    \label{fig:dataset_exp} \vspace{-0.5cm}
\end{figure*}

\label{sec:desc}
% In this section, we describe each of the eight datasets we constructed for the competition. The code to generate these datasets can be found on our GitHub\footnote{\hyperlink{https://github.com/RobGeada/cvpr-nas-datasets}{https://github.com/RobGeada/cvpr-nas-datasets}}.

The code to generate our datasets is available at our GitHub page\footnote{\href{https://github.com/Towers-D/NAS-Unseen-Datasets}{github.com/Towers-D/NAS-Unseen-Datasets}}. Each dataset comprises six NumPy files, X (images) and Y (labels) files for the training, validation, and testing sets, and a metadata file. This metadata file includes the shape of the training data, the ResNet-18\cite{he_deep_2016} benchmark result, and the number of classes in the data. Further details and examples are provided in the Appendices.

\subsection{AddNIST}
\label{desc:addnist}
AddNIST\cite{towers_addnist_2023}, is a Type-1 dataset of 70,000 images with an image shape of 3$\times$28$\times$28 (channels first). The training, validation, and testing split is 45,000, 15,000, and 10,000, respectively. Each colour channel is an image from MNIST \cite{yann_lecun_mnist_2005}. An example AddNIST image is depicted in \cref{fig:dataset_exp} (left), the larger image, shows the image as is, with the three channels laid on top of each other, and three images to the right show each channel more clearly.

AddNIST has twenty classes (0 - 19), with the class derived from the MNIST label of each channel, such that $l = (r + g + b) - 1$, where $l$ is the image label and $r, g,$ and $b$ are the respective MNIST labels of each colour channel.

AddNIST adds complexity to the MNIST data. In MNIST, the goal is to identify what numerical digit is seen within the image, while in AddNIST, not only does the model have to identify the digit in each channel, but it must also learn to sum up these values in a specific manner to identify the class correctly. In this manner, AddNIST was designed around the research question of whether NAS could ``figure out" a calculation was required and apply it.

While the task this dataset puts forward is still rather simple, which is expected for a Type 1 dataset, it requires high-level inference capabilities within the model, which would mean the NAS algorithm searching for the optimal architecture would have to consider the capacity of the model to encompass the required function. 

\subsection{Language}
\label{desc:lang}

\begin{figure}
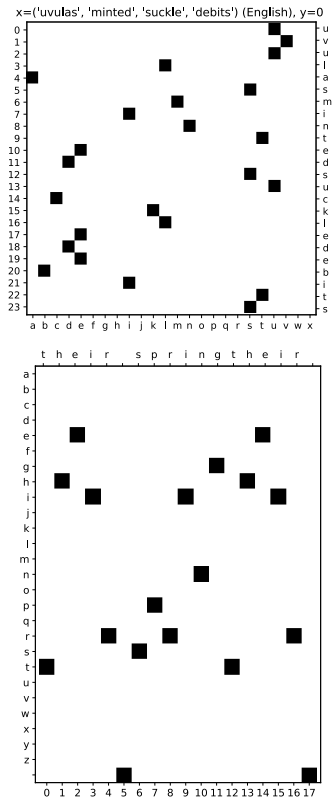

    \centering
    \includesvg[width=0.55\columnwidth]{Images/Language_single.svg}
    \includesvg[width=0.5\columnwidth]{Images/Gutenberg_single.svg}
    \caption{\textit{Top:} An example of a Language image with a readable axis included. The right-hand axis contains the four six-letter words ``Uvulas", ``Minted", ``Suckle", and ``Debits", which are all English words given the label 0. \textit{Bottom:} An example Gutenberg image, the words ``their", ``spring", and ``their" have been encoded from one of Shakespeare's works which give the label 4}
    \label{fig:langs} \vspace{-0.5cm}
\end{figure}

Using the open-source, public spell checker ASPELL, which can be found at \hyperlink{http://aspell.net}{aspell.net}, we created Language\cite{towers_language_2023}, a Type-2 dataset. 

Using dictionaries from 10 languages that use the Latin alphabet (English, Dutch, German, Spanish, French, Portuguese, Swedish, Zulu, Swahili, and Finnish), all six-letter words within each language are extracted, and any words with letters that use diacritics (such as é or ü) or include `y' or `z' are subsequently removed.

\begin{figure*}
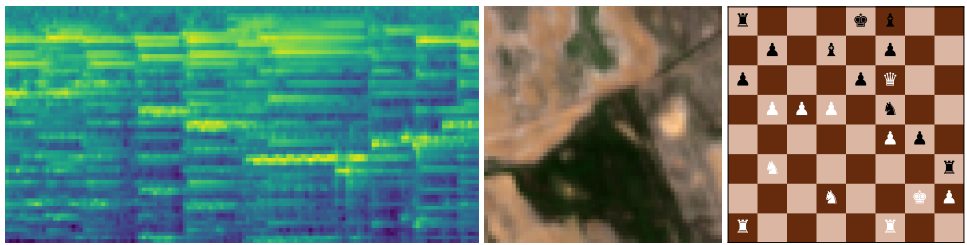

    \centering
    \includesvg[width=0.36\textwidth]{Images/Isabella_single.svg}
    \includesvg[width=0.18\textwidth]{Images/GeoClassing_single.svg}
    \includesvg[width=0.18\textwidth]{Images/Chesseract_single.svg}
    \caption{\textit{Left:} An example of an Isabella generated using a piece of music labelled as ``20th Century"(0). \textit{Middle} An example of the GeoClassing dataset showing a photo taken over Portugal (9). \textit{Right:} An example rendering of a board position in the Chesseract dataset wherein white goes on to eventually win and is thus given a label of White Wins (0)}
    \label{fig:new_sets} \vspace{-0.3cm}
\end{figure*}

Each image is generated by first selecting one of the ten languages. We then randomly select four words that we filter from the selected dictionary and concatenate them into a 24-character string. Since we eliminated letters with diacritics and y and z, we have a remaining alphabet of twenty-four letters. This allows for a 24$\times$24-grid encoding the string as a 1$\times$24$\times$24 image. Using the $y$-axis to denote the index of the string and the $x$-axis to denote the character, we construct the image so each black pixel denotes the letter used in that position. See \cref{fig:langs} (top) for an example. The training, validation, and testing split is 50,000, 10,000, and 10,000 images, respectively.

%These images require any ML model to perform linguistic analysis to identify which permutation of letters is more associated with a given language. 
The Language research question pertained to whether a linguistic character encoding contained enough information for DL to correctly identify the original language, requiring the model to perform linguistic analysis. To ensure no leakage between the training and testing data, when selecting a word for the training data, we ensure that this word does not appear in the validation or testing data and vice versa.

\subsection{MultNIST}
\label{desc:multnist}
MultNIST\cite{towers_multnist_2023} is a Type-1 dataset similar in concept to AddNIST (\cref{desc:addnist}), originating from the same research question. Like AddNIST, MultNIST images have a 3$\times$28$\times$28 channel first shape, and each channel is an image from the MNIST dataset \cite{yann_lecun_mnist_2005}. See \cref{fig:dataset_exp} (middle) for an example. The training, validation, and testing split is 50,000, 10,000, and 10,000 images, respectively.

MultNIST has only ten classes (0 - 9). The label of each MultNIST image is calculated such that $l = (r * g * b) \mod 10$ where $l$ is the image label and $r, g,$ and $b$ are the respective MNIST labels of each colour channel.

In AddNIST, the class was directly affected by how large the numbers were in each channel. The only way for an image to have a label of 0 was if one colour channel contained an MNIST with label 1 and the remaining channels contained MNIST images for 0. MultNIST introduces the additional complexity over MNIST by retaining a calculation but removes the bias of larger numbers to higher classes.

\subsection{CIFARTile}
\label{desc:ctile}
CIFARTile is a Type-1 dataset where each image in the CIFARTile\cite{towers_cifartile_2023} dataset is a compilation of four CIFAR-10 \cite{krizhevsky_learning_2009} images tiled together in a grid resulting in images which are of size 3$\times$64$\times$64. See \cref{fig:dataset_exp} (right) for an example. The training, validation, and testing split is 45,000, 15,000, and 10,000 images, respectively.

There are four classes (0 - 3), which represent the number of CIFAR-10 classes in each grid minus one. For example, a grid consisting of CIFAR-10 images with labels [\textit{horse}, \textit{horse}, \textit{frog}, \textit{car}] has three distinct classes, and thus a label of $3-1=2$. As such, CIFARTile is a Type-1 dataset; any human should be able to identify and solve this task.

CIFARTile requires a model to identify multiple CIFAR-10 classes simultaneously and determine which are similar and which differ. Whether models could adequately do this is the driving research question behind this dataset. \vspace{-0.2cm}

\begin{table*}[]
\centering
\begin{tabular}{l|rllrrrr|r}
\textbf{Dataset} & \multicolumn{1}{l}{\textbf{ResNet-18}} & \textbf{AlexNet} & \textbf{VGG16} & \multicolumn{1}{l}{\textbf{ConvNext}} & \multicolumn{1}{l}{\textbf{MNASNet}} & \multicolumn{1}{l}{\textbf{DenseNet}} & \multicolumn{1}{l|}{\textbf{ResNeXt}} & \multicolumn{1}{l}{\textbf{Random}} \\ \hline
AddNIST          & 92.08\%                                & \textbf{94.87\%} & 92.06\%        & 38.06\%                               & 90.51\%                              & 93.52\%                               & 91.42\%                               & 5\%                                 \\
Language         & 87.00\%                                & 85.71\%          & 84.54\%        & 83.40\%                               & 84.63\%                              & 84.57\%                               & \textbf{93.97\%}                      & 10\%                                \\
MultNIST         & 91.55\%                                & \textbf{94.01\%} & 90.43\%        & 64.20\%                               & 87.70\%                              & 92.81\%                               & 90.57\%                               & 10\%                                \\
CIFARTile        & 45.56\%                                & 48.88\% & 24.43\%        & 31.06\%                               & 48.49\%                              & \textbf{51.28\%}                              & 46.23\%                               & 25\%                                \\
Gutenberg        & 40.98\%                                & \textbf{45.53\%} & 44.00\%        & 31.93\%                               & 38.00\%                              & 43.28\%                               & 40.30\%                               & $16.\dot{6}$\%                              \\
Isabella         & 62.02\%                                & 61.37\%          & 58.13\%        & 57.18\%                               & 60.69\%                              & \textbf{63.27\%}                      & 60.46\%                               & 25\%                                \\
GeoClassing      & 80.33\%                                & 92.49\%          & 83.67\%        & 72.76\%                               & 86.00\%                              & \textbf{94.21\%}                      & 89.99\%                               & 10\%                                \\
Chesseract       & 57.83\%                                & 57.45\%          & 55.69\%        & 52.74\%                               & 56.26\%                              & \textbf{59.60\%}                      & 55.15\%                               & $33.\dot{3}$\%                              \\ 
\end{tabular}
% \caption{Experiments on ResNet-18 \cite{he_deep_2016}, AlexNet \cite{krizhevsky_imagenet_2012}, VGG16 \cite{k_simonyan_very_2014}, ConvNext \cite{zhuang_liu_convnet_2022}, MnasNet \cite{tan_mnasnet_2019}, DenseNet \cite{gao_huang_densely_2017}, ResNeXt \cite{xie_aggregated_2017} using the presented datasets.}
\caption{Experiments on commonly-used CNN-based classification models \cite{he_deep_2016, krizhevsky_imagenet_2012, k_simonyan_very_2014, zhuang_liu_convnet_2022, tan_mnasnet_2019, gao_huang_densely_2017, xie_aggregated_2017} using the presented datasets.}
\label{tab:cnn-res} \vspace{-0.5cm}
\end{table*}

\subsection{Gutenberg}
\label{desc:gutenberg}
Gutenberg\cite{towers_gutenberg_2023} is a Type-2 dataset named after the source of the data, Project Gutenberg, found at \href{https://www.gutenberg.org}{www.gutenberg.org}, which provides free ebooks of literary works that are no longer under US copyright protection. We selected six authors (Thomas Aquinas, Confucius, Hawthorne, Plato, Shakespeare, and Tolstoy) and downloaded several books from each author -- see the Appendices for further details and examples.

The selected works are English translations that represent a variety of cultures and time periods. We perform basic text preprocessing, including removing punctuation, converting letters with diacritics to the base letter, and removing ``structure" words (e.g., `Chapter', `Scene', `Prologue'). We then extracted consecutive sequences of three words between 3 and 6 letters long. In each sequence, the three words were padded up to 6 characters with spaces. Then, the three words were concatenated together to produce an 18-character string. These strings were used as the base for image creation. Training, test, and validation sequences were chosen such that there was no overlap between any sequence across any data split.

Similar to the Language dataset (\cref{desc:lang}), we converted these strings into images. On a 27$\times$18 grid, we created a mapping of each character in the string. See \cref{fig:langs} (bottom) for an example. The $x$-axis represents the position in the string, and the $y$-axis represents the alphabetical letter (or space) located at that position. This results in the shape being 1$\times$27$\times$18. The task is then to predict which of the original six authors the original word sequence came from. The training, validation, and testing split is 45,000, 15,000, and 6,000, respectively, to prevent participants from exploiting the standard distribution we used for the other datasets.

The research question behind this dataset was whether specific spatial patterns of letters were sufficient information for a neural network to identify the authors from which they originated, which seems impossible from face value. 

\subsection{Isabella}
\label{desc:isabella}
Isabella\footnote{Code to generate Isabella is available at \href{https://github.com/Towers-D/NAS-Unseen-Datasets}{github.com/Towers-D/NAS-Unseen-Datasets}.} is a Type-2 dataset that uses musical recordings from the Isabella Stewart Gardner Museum, Boston.

The four classes of this dataset refer to the era of composition (Baroque, Classical, Romantic, and 20th Century) attributed by the museum. The recordings are split into five-second snippets converted into 64-band spectrograms, creating a dataset of 1$\times$64$\times$128 shaped images. See \cref{fig:new_sets} (left) for an example. The training, validation, and testing split is 50,000, 10,000, and 10,000 images, respectively.

The task for the models is to predict the era of composition from the spectrogram. No recording used in a training snippet appeared in the validation or test sets. The research question behind this dataset was to explore whether era-defining characteristics appear at the spectrogram level and whether CNNs could reliably identify such patterns. 

\begin{table*}[]
\centering
\begin{tabular}{lrrrrrr}
                                      & \multicolumn{1}{l}{}                  & \multicolumn{1}{l}{}               & \multicolumn{1}{l}{}                     & \multicolumn{2}{c}{\textbf{Random Search}}                                & \multicolumn{1}{l}{}                \\
\multicolumn{1}{l|}{\textbf{Dataset}} & \multicolumn{1}{l}{\textbf{PC-DARTS}} & \multicolumn{1}{l}{\textbf{DrNAS}} & \multicolumn{1}{l|}{\textbf{Bonsai-Net}} & \multicolumn{1}{l}{\textbf{DARTS}} & \multicolumn{1}{l|}{\textbf{Bonsai}} & \multicolumn{1}{l}{\textbf{Random}} \\ \hline
\multicolumn{1}{l|}{Addnist}          & 96.60\%                               & 97.06\%                            & \multicolumn{1}{r|}{\textbf{97.91\%}}    & 97.07\%                            & \multicolumn{1}{r|}{34.17\%}         & 5\%                                 \\
\multicolumn{1}{l|}{Language}         & \textbf{90.12\%}                      & 88.55\%                            & \multicolumn{1}{r|}{87.65\%}             & 90.12\%                            & \multicolumn{1}{r|}{76.83\%}         & 10\%                                \\
\multicolumn{1}{l|}{MultNIST}         & 96.68\%                               & \textbf{98.10\%}                   & \multicolumn{1}{r|}{97.17\%}             & 96.55\%                            & \multicolumn{1}{r|}{39.76\%}         & 10\%                                \\
\multicolumn{1}{l|}{CIFARTile}        & \textbf{92.28\%}                      & 81.08\%                            & \multicolumn{1}{r|}{91.47\%}                   & 90.74\%                                  & \multicolumn{1}{r|}{24.76\%}         & 25\%                                \\
\multicolumn{1}{l|}{Gutenberg}        & \textbf{49.12\%}                      & 46.62\%                            & \multicolumn{1}{r|}{48.57\%}             & 47.72\%                                  & \multicolumn{1}{r|}{29.00\%}         & $16.\dot{6}$\%                              \\
\multicolumn{1}{l|}{Isabella}         & \textbf{65.77\%}                                     & 64.53\%                   & \multicolumn{1}{r|}{64.08\%}                   & 66.35\%                                  & \multicolumn{1}{r|}{58.53\%}         & 25\%                                \\
\multicolumn{1}{l|}{GeoClassing}      & 94.61\%                               & \textbf{96.03\%}                   & \multicolumn{1}{r|}{95.66\%}                   & 95.54\%                                  & \multicolumn{1}{r|}{63.56\%}         & 10\%                                \\
\multicolumn{1}{l|}{Chesseract}       & 57.20\%                               & 58.24\%                   & \multicolumn{1}{r|}{\textbf{60.76\%}}                   & 59.16\%                            & \multicolumn{1}{r|}{68.83\%}         & $33.\dot{3}$\%                             
\end{tabular}
\caption{Experimental results on PC-DARTS\cite{xu_pc-darts_2019}, DrNAS\cite{chen_drnas_2020}, Bonsai-Net \cite{geada_bonsai-net_2020} as well as random search.}
\label{tab:nas-res} \vspace{-0.5cm}
\end{table*}

\subsection{GeoClassing}
\label{desc:GeoClassing}
GeoClassing\cite{towers_geoclassing_2023}, known as Sadie in the competition, is a Type-2 dataset that takes advantage of the BigEarthNet dataset \cite{gencer_sumbul_bigearthnet_2019} as a foundation. The BigEarthNet dataset, as originally published, consists of satellite photography with ground-cover classification labels, i.e., ``airports" or ``vineyards". An example image is shown in \cref{fig:new_sets}
 (middle).
 
The GeoClassing dataset uses these images, but instead of using the given ground-cover labels, we identified the European Space Agency Sentinel patch from which the BigEarthNet image was sourced. We cross-referenced the coordinates of that patch within a map of Europe to identify the corresponding country depicted in the image. This gave us ten classes (Austria, Belgium, Finland, Ireland, Kosovo, Lithuania, Luxembourg, Portugal, Serbia, and Switzerland).

Each image has a shape of 3$\times$60$\times$60, and the training, validation, and testing split contains 43,821, 8,758, and 8,751 images, respectively. The research question is whether NAS can find models that identify countries from differences in topology and ground coverage, which, without specific knowledge of geography and vegetation type, would be extremely difficult for a human to do manually.

\subsection{Chesseract}
\label{desc:chesseract}
Chesseract\cite{towers_chesseract_2023}, a Type-1 dataset wherein we accessed public chess games from eight grandmasters (Bobby Fischer, Garry Kasparov, Magnus Carlsen, Viswanathan Anand, Hikaru Nakamura, Anatoly Karpov, Fabiano Caruana, and Mikhail Tal) extracting the final 15\% of board states. These positions were then one-hot encoded by piece type and colour, creating a 12$\times$8$\times$8 image. As a 12-channel image is hard to render, we provided a flattened 2D image of one of the chess boards we used \cref{fig:new_sets} (right). A 3D visualisation can be seen in the Appendices.

The training, validation, and testing split is 49,998, 9,999, and 9,999 images, respectively. Each image in the Chesseract dataset is one of three classes (White wins, Draw, Black wins). No individual positions from the same game appeared across the data splits. This dataset requires the model to identify the game's final result from the position. Chesseract presented a problem for some participants of the challenge. This was because Chesseract uses twelve instead of the usual 1 or 3 channels applied for image datasets, which often caused errors for hardcoded algorithms that only accepted 1 or 3 channel dimensions.

Machine learning is already prominent in chess, with competing learning-based chess engines that beat grandmaster players, such as AlphaZero\cite{david_silver_mastering_2017}. Our research question for this dataset was based on the understanding that any position can be analysed to show whether black or white has the advantage. Is NAS able to develop a DL network to identify this from just an encoding of the board position?

% \subsection{Fashion-MNIST}
% During the first competition, we included Fashion-MNIST as a Type-1 dataset \cite{xiao_fashion-mnist_2017}, with the code name Fabian, as one of the three datasets given to participants to test their algorithms. Fashion-MNIST is similar to the more popular MNIST \cite{lecun_mnist_2010} dataset, using articles of clothing instead of numbers.

% The Fashion-MNIST dataset is another Type-1 dataset \cite{xiao_han_fashion-mnist_2017}, similar to the more popular MNIST~\cite{yann_lecun_mnist_2005} dataset, but using articles of clothing. We included Fashion-MNIST in the first challenge even though it is not an `unseen' dataset as it allowed NAS developers to work with known datasets -- which NAS approaches should also support.

\section{Baseline Experimentation}
\label{base-ref}
Here, we provide some baseline results for our datasets to motivate future NAS experimentation. We provide a set of baseline results across several well-known CNNs and three NAS methods across two search spaces.

\subsection{CNN Experiments}
\label{Res:CNNS}
To generate baseline results of our datasets for future work comparison, we have performed experiments using commonly used off-the-shelf CNNs, which can be seen in \cref{tab:cnn-res}. In all our CNN experiments, we have followed a similar experimental setup. We use stochastic gradient descent as our optimiser with an initial learning rate of 0.01, momentum of 0.9, and a weight decay of $3 \times 10^{-4}$. We have used a Cosine Annealing Learning rate with the max number of iterations equaling 64, the number of epochs. Cross Entropy is used as the Loss function.

ResNet-18 is used as the baseline throughout our competition, and any NAS method would be expected to easily outperform this architecture. As you can see in \cref{tab:cnn-res}, ResNet-18 produces competitive results across our datasets, though it does not excel on any particular dataset. ResNet-18 particularly struggles with the GeoClassing dataset compared to the others, only beating ConvNext in our trials. %This is most likely due to the fact that ResNet-18, with its relatively shallow architecture, does not have sufficient capacity to capture and discriminate between the complex spatial patterns within GeoClassing effectively.

AlexNet \cite{krizhevsky_imagenet_2012} shows decent performance across all datasets, achieving the highest accuracy on three of the eight datasets (AddNIST, MultNIST, and Gutenberg). AddNIST and MultNIST are similar, both being derived from MNIST. This may suggest that AlexNet is particularly good at working with MNIST Images. AlexNet captures spatial hierarchies in images well and is thus suitable for the clear, structured layouts of MNIST digits. VGG16 \cite{k_simonyan_very_2014}, on the other hand, does not achieve particularly impressive levels of accuracy on any of the datasets, though it does not perform badly, either. The only exception is the CIFARTile dataset, where VGG suffers and only achieves roughly the same accuracy as random chance, as seen in \cref{tab:cnn-res}.

ConvNext \cite{zhuang_liu_convnet_2022} performs poorly across our datasets, most likely since our datasets may not exhibit the specific spatial complexity that ConvNext is optimised for. However, it is important to note that for purposes of generating our CNN baselines, we have kept the experimental setup the same across CNNs. Thus, hyperparameter tuning may be beneficial to ConvNext. We have used the base version of ConvNext in our experiments. 

We also perform the same experiments on MnasNet \cite{tan_mnasnet_2019} using a depth multiplier of 1.0. Despite being a lightweight architecture designed for mobile efficiency, MnasNet achieves competitive results across our datasets. It has strong generalisation capabilities since it uses a reinforcement learning approach to automate architecture design, thus performing well on unseen datasets \cite{tan_mnasnet_2019}.

DenseNet \cite{gao_huang_densely_2017} performs significantly well on the latter three datasets (Isabella, GeoClassing, and Chesseract) as well as CIFARTile, achieving the highest accuracy of the CNNs over these datasets. With the exception of Chesseract, the datasets that DenseNet excels on are the most memory-intensive datasets. This is due to the DenseNet mechanism allowing efficient re-use of features through dense connectivity and ensuring maximal information flow between layers in the network \cite{gao_huang_densely_2017}. For memory-intensive datasets, which often contain complex and rich information, this feature reuse allows DenseNet to exploit the data more thoroughly and efficiently, leading to better performance. The DenseNet-161 architecture is used in our experiments.

Finally, ResNeXt \cite{xie_aggregated_2017} performs fairly well across our datasets, but especially well on the Language Dataset, scoring almost 7\% higher than the other methods. The particular version of ResNeXt we used is ResNeXt-50 (32 x 4d), which has a high level of cardinality (the size of the set of transformations) \cite{xie_aggregated_2017}. This high cardinality allows ResNeXt to learn more complex and diverse representations, which is crucial for distinguishing subtle patterns in the Language Dataset, such as the frequency and arrangement of characters encoded in images.

Our analysis across these CNN architectures reveals distinct performance trends tied to the nature of each dataset. ResNet-18 serves as a versatile baseline, while AlexNet is especially capable of handling structured patterns. VGG16's mixed results and ConvNext's overall underperformance emphasise the need within conventional deep learning for matching architectural strengths to dataset characteristics, which is not desirable for evaluating generalisation capabilities in general and NAS algorithms in particular. DenseNet and ResNeXt are good at handling memory-intensive datasets and those with nuanced patterns, respectively. These insights confirm that the effectiveness of a CNN architecture is highly contingent on the specific demands of the dataset, which underscores the need for unseen datasets for effective evaluation of NAS methods.

\subsection{NAS Experiments}
In addition to the experiments performed on the CNNs, we have also applied the NAS methods of PC-DARTS \cite{xu_pc-darts_2019}, DrNAS \cite{chen_drnas_2020} (DARTS-space), and Bonsai-Net \cite{geada_bonsai-net_2020} to our datasets. Furthermore, as a baseline of comparison, we perform a random search in both the DARTS and Bonsai search spaces, using the provided methods from PC-DARTS and Bonsai-NET. These results can be found in \cref{tab:nas-res}.

During our random search experiments using the DARTS search space, it is noted that random search often outperforms either PC-DARTS or DrNAS and outperforms both models on AddNIST, Chesseract and Isabella, tying PC-DARTS on Language while outperforming DrNAS. These results support findings by Yu et al.\cite{yu_evaluating_2019} that searches using these search spaces often do not significantly outperform random search. %MultNIST is the only dataset where both PC-DARTS and DrNAS outperform random search.

We can see in \cref{tab:nas-res} that PC-DARTS\cite{xu_pc-darts_2019} performs well, scoring the highest accuracies over four of the eight datasets and gives competitive results over the others. DrNAS\cite{chen_drnas_2020}, however, scores the highest on two of the eight datasets but struggles with CIFARTile, where it scores 10\% lower than the other methods. We used the DARTS search space during our DrNAS experiments; while DrNAS is also available to work on the NAS-Bench-201\cite{dong_nas-bench-201_2020} search space, it employs the NAS-Bench API to look at the performance of architectures fully trained on CIFAR-10, which would not reflect the performance these architectures found on our datasets. In most cases, Bonsai-Net \cite{geada_bonsai-net_2020} outperforms random search, with the exception of Chesseract. It returns the best results on AddNIST and Chesseract (though it is significantly beaten by random search on the latter dataset).

The varied performance of these NAS methods across our datasets highlights their ability to challenge and assess the adaptability and effectiveness of different architecture search strategies. This underscores the strengths of our datasets in providing a comprehensive benchmark for evaluating NAS methods' capability to generalise across a diverse range of tasks and complexities.

\section{Discussion}

\begin{table}[]
\centering
\begin{tabular}{l|rrr}
Dataset     & \multicolumn{1}{l}{CNN} & \multicolumn{1}{l}{NAS} & \multicolumn{1}{l}{Competition} \\ \hline
AddNIST     & 94.87\%                 & \textbf{97.91\%}        & 95.06\%                         \\
Language    & \textbf{93.97\%}        & 90.12\%                 & 89.71\%                         \\
MultNIST    & 94.01\%                 & \textbf{98.10\%}        & 95.45\%                         \\
CIFARTile   & 51.28\%                 & \textbf{92.28\%}        & 73.08\%                         \\
Gutenberg   & 45.53\%                 & 49.12\%                 & \textbf{50.85\%}                \\
Isabella    & 63.27\%                 & \textbf{65.77\%}        & 61.42\%                         \\
GeoClassing & 94.21\%                 & 96.03\%                 & \textbf{96.08\%}                \\
Chesseract  & 59.60\%                 & 60.76\%                   & \textbf{62.98\%}               
\end{tabular}
\caption{Results on each dataset from our CNN and NAS experiments, and Competition Submissions}
\label{tab:res-comp} \vspace{-0.7cm}
\end{table}

\cref{tab:res-comp} shows the best-reported performance of the CNN and NAS experiments on each dataset, as well as the best result found by competition participants; for further information, see the Appendices. From these results, we can see that NAS found the best model for seven datasets, supporting the usage of NAS methods in finding good models for given datasets. Three of these architectures were found by competition submissions, NAS methods designed for unseen data; given the competition's time restrictions this may suggest that methods designed in this way can generalise to other datasets better.

%Furthermore, three of these datasets were found by NAS methods designed without knowing the evaluation data, although they were limited by time.

It is important to mention that we did not test the deepest versions of these CNN architectures or perform hyperparameter optimisation. Similarly, while we allowed the traditional NAS methods to run in their preset configuration, the competition participants were only given twenty-four hours to evaluate across three datasets. However, even with these limitations, CNNs and NAS architectures from the competition outperformed traditional NAS methods across several datasets, reinforcing our belief that current NAS methods do not necessarily generalise properly to unseen datasets.

%It is interesting to see how the models rank compared to each other across the different datasets. While AlexNet was the best on five datasets, ResNet-18 outperformed in the other three. Although AlexNet performs well across the datasets we experimented with, it is not a one-size-fits-all solution. This shows that a single neural architecture cannot be used to achieve optimum performance on every problem, and therefore, benchmarking the method on a few datasets will not indicate the performance across the domain. We believe this is also true of NAS methods, and evaluating based on a few datasets does not indicate a good solution. 

%It is important to state that the standard CNNs we tested were not very deep. There are deeper versions of both ResNet and VGG, which may perform better than their shallower counterparts that we experimented with. We also only tested the standard CNNs with a single experimental setup, while the models found in the competition were not limited to this. Hyperparameter optimisation could alter our results. Furthermore, we report each standard CNN's results separately, while we only report the best NAS model for each dataset from multiple competitors.

\section{Rights and Reproduction}
These datasets have been created under the licence agreements of the original data. Where available, we have made the datasets publicly accessible under an \href{https://creativecommons.org/licenses/by/4.0/}{CC BY 4.0 Licence}. The Isabella dataset uses data from the Isabella Stewart Gardner Museum, which withholds the right to share modifications to the music they have made available. Instead of providing the dataset, we provide a Python script to convert music files obtained from the Museum\footnote{\href{https://www.gardnermuseum.org/experience/music}{Available here https://www.gardnermuseum.org/experience/music}} into the format of the competition dataset on our GitHub \footnote{\href{https://github.com/Towers-D/NAS-Unseen-Datasets}{github.com/Towers-D/NAS-Unseen-Datasets}}.

\section{Conclusion}
Machine learning is a valuable and fast-growing tool that is quickly becoming part of the tools we use daily. For small businesses to stay competitive, they need to be able to easily include machine learning techniques in their products or business strategy. NAS promises to be a tool that removes the cost of an expert's time and knowledge to create bespoke neural networks while remaining competitive. NAS is currently evaluated primarily on a few benchmark datasets and developed based on these datasets as well. This does not reveal how generalisable NAS methods are when given a dataset, especially when the problem is distinctly different.

This paper introduces eight new datasets to be used when testing NAS methods. These datasets represent problems that are either simple or difficult for humans to solve and provide difficulty outside of normal image classification. While we believe using these datasets will improve the generalisability of NAS methods, the problem of not knowing whether NAS is good in general or only good on benchmark datasets is a rolling problem. Simply including these datasets for upcoming NAS works will mean they become part of the datasets developed in mind. To solve this problem, more datasets will be continually needed.

In the future, we seek to develop further datasets to enable people to work on `unseen' data for NAS. We will also evaluate these datasets against pre-existing, and NAS approaches we are developing.

For a video overview of the paper, please follow this link: \href{https://youtu.be/YdYHdxNZUIw}{youtu.be/YdYHdxNZUIw}

%%%%%%%%% REFERENCES
{\small
\bibliographystyle{ieee_fullname}
\bibliography{datasets}
}

\end{document}

\onecolumn
%\appendix
%\section{Dataset Examples and Licensing}

%This supplementary material provides additional examples and details on the datasets and licensing.

\section{AddNIST}
\label{App:ANIST}

The AddNIST dataset contains twenty classes [0,19] such that $l = (r + g + b) - 1$, where $l$ is the image label and $r, g,$ and $b$ are the respective MNIST labels of each colour channel, sampled from a list of combinations of integers 0-9. Note we exclude the case $r=0, g=0, b=0$. At first glance, the images look very similar to each other. Unlike MNIST or CIFAR-10, which are quickly identifiable for a human, AddNIST can be hard for a human to classify, as some images have MNIST numbers, which are hard to identify without being separated.

\begin{figure}[h]
    \centering
    \includesvg[width=\textwidth]{AppendicesImgs/AddNIST/AddNIST-frequency.svg}
    \caption{The count of MNIST class permutations for each AddNIST class.}
    \label{fig:addfreq}
\end{figure}

AddNIST suffers from a bias with some classes, especially images that sum towards the middle of the class range, where more permutations of numbers can have the same label. This frequency distribution can be seen in \cref{fig:addfreq}. 

This means the models can ``cheat" as low channel numbers mean lower labels and large channel numbers mean larger labels. The class label 0 can only be achieved from three MNIST label combinations ((0,0,1), (0,1,0) and (1,0,0)) whilst the class label 19 has 36 possible MNIST label combinations (for example, (6, 6, 8) and (9, 9, 2)). Furthermore, if one of the channels is a 9, then eight labels can already be discarded. 

The data generation procedure for this dataset did try to account for this, however. For each class between 0 and 19, all possible number triples that produced that class were enumerated. For example, the possible triples for class 0 are [(0, 0, 1), (0, 1, 0), and (1, 0, 0)]. After each triple for each class is determined, the total set of triples is divided into training, testing, and validation sets to ensure that the triple (a,b,c) only appears in one of the three data splits. Images were then generated from these sets of triples, such that the total number of each class in each split was equal. The intention here was to roughly balance the number of different combinations that produced a specific class while ensuring that the models could not simply overfit to the specific digit combination. 

\cref{App-fig:AddNIST} shows some sample renderings of the AddNIST dataset, one for each class, alongside the channel breakdown.

\begin{figure}[h]
    \includesvg[width=\textwidth]{AppendicesImgs/AddNIST/AddNIST-row-1.svg}
    
    \includesvg[width=\textwidth]{AppendicesImgs/AddNIST/AddNIST-row-2.svg}
    
    \includesvg[width=\textwidth]{AppendicesImgs/AddNIST/AddNIST-row-3.svg}
    
    \includesvg[width=\textwidth]{AppendicesImgs/AddNIST/AddNIST-row-4.svg}
    
    \includesvg[width=\textwidth]{AppendicesImgs/AddNIST/AddNIST-row-5.svg}
\end{figure}
\begin{figure}
    \includesvg[width=\textwidth]{AppendicesImgs/AddNIST/AddNIST-row-6.svg}
    
    \includesvg[width=\textwidth]{AppendicesImgs/AddNIST/AddNIST-row-7.svg}
    \caption{Example renderings of each AddNIST dataset class alongside the individual MNIST images that make full AddNIST image.}
    \label{App-fig:AddNIST}
\end{figure}

\section{Language}
\label{App:Lang}

The Language dataset contains ten classes of language-encoded images. Each class has an associated language with six-letter words extracted. These words have been encoded using a character-to-space mapping. The possible languages that a Language image can be are English, Dutch, German, Spanish, French, Portuguese, Swedish, Zulu, Swahili, and Finnish.

Each image has four randomly selected words from the associated language. These words do not include diacritics (letters such as `é' or `ü') or the letters `y' or `z'. The latter exclusion had two motivations: the first, to produce a square 24x24 encoding (4, 6-letter words that use 24 possible letters); the second, as a dimension of 26 might hint at an alphabetical encoding, the choice of 24 letters obscures this. 

The words are appended together to form a twenty-four-character-long string. We then encode this string into an image through a grid representation. The $y$-axis denotes the index of the character in the string, while the $x$-axis shows the letter. 

\cref{App-fig:lang} renderes some examples of the Language dataset, showing a sample from each class. Above the image, we display the randomly sampled words and their associated language label. To aid in understanding the letter from the 24-character string is shown on the right-hand side of each `image' -- you can read the words downwards. %To the right, the string is shown on the same horizontal line as its associated point in the grid. 

\begin{figure}[h]
    \centering
    \includesvg[width=\textwidth]{AppendicesImgs/Language/Language-row-1.svg}
    \includesvg[width=\textwidth]{AppendicesImgs/Language/Language-row-2.svg}

    \includesvg[width=\textwidth]{AppendicesImgs/Language/Language-row-3.svg}

    \includesvg[width=\textwidth]{AppendicesImgs/Language/Language-row-4.svg}
    \caption{Example renderings of the Language dataset, each sample includes the language and chosen words above, the index position on the left axis, and the possible letters in alphabetic order along the bottom.}
    \label{App-fig:lang}
\end{figure}

\section{MultNIST}
\label{App:MuNIST}
The MultNIST dataset is designed in a similar way to the AddNIST dataset \ref{App:ANIST}. MultNIST differs from AddNIST in the formula used to calculate the dataset labels, using multiplication and taking the modulus (base 10) of the result instead of using addition. I.e., $l = (r * g * b) \mod 10$ where $l$ is the image label and $r, g,$ and $b$ are the respective MNIST labels for each colour channel.

Since the modulus operation is performed after the multiplication, MultNIST only contains ten classes (0 - 9), which indicates the last digit of the multiplication (base 10). Due to this, MultNIST does not suffer from the numerical label bias that AddNIST suffers from. This is due to the modulus operation, which was chosen for the specific purpose of diversifying the possible digit combinations that could produce each class; without it, class 0 would only consist of combinations that contained at least one zero. With the modulo, all triples that contain any permutation of $(5, 2*a, b) \forall a, b \in [0, 9]$ are also class 0, for example. 

For each class, we have rendered a sample image in \cref{App-fig:mult}, with the channel breakdown shown above and alongside the main image.
\begin{figure}[h]
    \includesvg[width=\textwidth]{AppendicesImgs/MultNIST/MultNIST-row-1.svg}

    \includesvg[width=\textwidth]{AppendicesImgs/MultNIST/MultNIST-row-2.svg}

    \includesvg[width=\textwidth]{AppendicesImgs/MultNIST/MultNIST-row-3.svg}

    \includesvg[width=\textwidth]{AppendicesImgs/MultNIST/MultNIST-row-4.svg}
    
    \caption{Example renderings of each MultNIST class alongside the MNIST images that make the full MultNIST.}
    \label{App-fig:mult}
\end{figure}

\section{CIFARTile}
\label{App:CTILE}
The CIFARTile dataset contains four classes representing the number of CIFAR-10 classes that the sub-images are associated with, minus one (for zero-indexing). The images are generated by randomly choosing the number of classes to include in the image. We then randomly choose which classes are to be included, creating a list containing the necessary duplicates. We then use the selected classes, to randomly sample CIFAR-10 images of the selected class.

Once we have selected the four sub-images, we perform random cropping, pad the images back up to 28x28, and then concatenate the images into a 56x56 image.

Sample renderings of each CIFARTile class can be seen in \cref{App-fig:ctile}.

\begin{figure}[h]
    \includesvg[width=\textwidth]{AppendicesImgs/CIFARTile/CIFARTile-row-1.svg}

    \includesvg[width=\textwidth]{AppendicesImgs/CIFARTile/CIFARTile-row-2.svg}
    
    \label{Example renderings of the CIFARTile dataset. The CIFAR-10 classes that make up each image are shown above.}
    \label{App-fig:ctile}
\end{figure}

\section{Gutenberg}
\label{App:Gute}
The Gutenberg dataset consists of six classes, each representing an author of a famous literary work. Here, each e-book from Project Gutenberg was first processed to remove non-body text, such as headers, licenses, tables-of-contents, and book-structural words (such as Chapter, Act, Prologue, Scene, etc). Then, all three-word sequences (\textit{phrases}) from each book that exclusively contained 3-6 Latin-lettered words were extracted. For example, a possible phrase for Shakespeare might be ``art thou romeo". All words were then padded to six characters via underscores: \texttt{art\char`_\char`_\char`_,thou\char`_\char`_,romeo\char`_}. Any phrase that appeared across multiple authors was removed. Similar to the Language Dataset, we performed a character mapping into a grid. Unlike Language, Gutenberg includes all 26 Latin characters plus underscores, thus creating images of size 27x18. 

The literary works used were accessed from Project Gutenberg\footnote{\href{https://www.gutenberg.org}{https://www.gutenberg.org}}. Please note that the works in Project Gutenberg are no longer under US copyright. This may still be protected under copyright in other countries. Please check before downloading our dataset or work from Project Gutenberg.

We accessed works from Thomas Aquinas (Summa I-II, Summa Theologica, Part III, and On Prayer and the Contemplative Life), Confucius (The Sayings of Confucius and The Wisdom of Confucius), Hawthorne (The Scarlet Letter and The House of the Seven Gables), Plato (The Republic, Symposium, and Laws), Shakespeare (Romeo and Juliet, Macbeth, Merchant of Venice, and King lear), and Tolstoy (War and Peace, and Anna Karenina). 

\cref{App-fig:gute} shows example renderings of the dataset.

\begin{figure}[h]
    \includesvg[width=\textwidth]{AppendicesImgs/Gutenberg/Gutenberg-row-1.svg}

    \includesvg[width=\textwidth]{AppendicesImgs/Gutenberg/Gutenberg-row-2.svg}
    
    \caption{Example renderings of the Gutenberg dataset. The $x$-axis shows the index position in the string and the $y$-axis shows the character at the corresponding position.}
    \label{App-fig:gute}
\end{figure}

\section{Isabella}
\label{App:Isabella}
The Isabella dataset is constructed by creating spectrograms of non-overlapping five-second audio clips, giving them a label based on the era of music from which they were composed. Here, songs were split into non-overlapping 5-second chunks. The sum absolute amplitude of the audio signal was used as a threshold, to filter out 5-second windows that were majority silent. Then, each chunk was converted into a 128-band spectrogram with 256 time buckets. The training, test, and validation sets were chosen such that no recording appears across multiple sets to ensure models do not overfit to artefacts in the recording environment (for example, specific microphone characteristics or room reverberations).

The audio used to create the original dataset was taken from the online library of the Isabella Stewart Gardner Museum, Boston. The music can be accessed from the museum under a \href{https://creativecommons.org/licenses/by-nc-nd/4.0/}{Creative Commons license (CC BY-NC-ND 4.0)} at \href{https://www.gardnermuseum.org/experience/music}{www.gardnermuseum.org/experience/music}. While this licence allows us to create and use the dataset, it does not permit us to distribute it. Instead, the code used to generate the dataset is available on our GitHub\footnote{\href{https://github.com/Towers-D/NAS-Unseen-Datasets}{github.com/Towers-D/NAS-Unseen-Datasets}}.

The labels we have used are based off of the classification of each piece by the museum - the classes for Isabella are Romantic, 20th Century, Baroque, and Classical.

\cref{App-fig:bella} shows example renderings of the Isabella dataset.

\begin{figure*}
    \centering
    \includesvg[width=\textwidth]{AppendicesImgs/Isabella/Isabella-row-1.svg}
    \includesvg[width=\textwidth]{AppendicesImgs/Isabella/Isabella-row-2.svg}
    
    \caption{Rendered examples of the Isabella dataset.}
    \label{App-fig:bella}
\end{figure*}

\section{GeoClassing}
\label{App:GeoClassing}
The GeoClassing data was created using the BigEarthNet dataset using the following licence \href{https://bigearth.net/downloads/documents/License.pdf}{https://bigearth.net/downloads/documents/License.pdf} and using the data available from the European Space Agency at \href{https://sentinel.esa.int/web/sentinel/sentinel-data-access}{https://sentinel.esa.int/web/sentinel/sentinel-data-access}.

Labels were given to the original data representing the land-cover type of the image, such as forest or ``urban fabric". However, each image is one of ten European countries; thus, we use country-of-origin as the class label. These countries are Austria, Belgium, Finland, Ireland, Kosovo, Lithuania, Luxembourg, Portugal, Serbia, and Switzerland. This was retrieved by matching each BigEarth image to the Sentinel patch it originates from, then cross-referencing the coordinates of the source Sentinel images against a map of Europe. A total of 56 Sentinel images are represented in the dataset. The sum total pixel value is used as a threshold to filter out images that were too dark or too light, thus removing images obscured by clouds, snow-covered, taken at night, etc. Furthermore, any image assigned the BigEarth label of ``Sea and ocean" or ``Water bodies" was removed, as those would unlikely contain country-identifying information.

This dataset can be difficult for a human to solve by eye, as seen in \cref{App-fig:geo} in our sample renderings. The samples of Ireland, Lithuania, Austria, and Luxembourg all show images of farmland, which would be hard to differentiate from sight alone without in-depth knowledge of European vegetation. Some images may be more obvious, for example, the presence of snow might rule out Portugal.

\begin{figure*}
    \centering
    \includesvg[width = 0.85\textwidth]{AppendicesImgs/GeoClassing/GeoClassing-row-1.svg}

    \includesvg[width = 0.85\textwidth]{AppendicesImgs/GeoClassing/GeoClassing-row-2.svg}

    \includesvg[width = 0.85\textwidth]{AppendicesImgs/GeoClassing/GeoClassing-row-3.svg}

    \includesvg[width = 0.85\textwidth]{AppendicesImgs/GeoClassing/GeoClassing-row-4.svg}
    
    \caption{Rendered examples of the GeoClassing Dataset.}
    \label{App-fig:geo}
\end{figure*}

\section{Chesseract}
\label{App:Chesseract}
The games used to create Chesseract were downloaded from Chess.com, using their publicly accessible tools. We downloaded every listed game attributed to one of eight grandmasters. Games that were played between two grandmasters in our originating set were assigned to just one of the players, thus de-duplicating them. For each game, we extracted \textit{board states}, i.e., the positioning of all pieces after $n$ moves. For a game with $N$ total board states, the states from $[N*.85, N)$ were selected to correspond to near-endgame/endgame board states while excluding the very final state of the board. Games were then split into training, testing, and validation sets to ensure that no two board states from the same game appeared in different data splits. Each individual data split may contain multiple states from the same game, however. 

The board states are encoded into a 12-channel format. We use the twelve channels to one-hot encode each piece type (pawn, knight, rook, bishop, queen king -- all both black and white). While this can be visualised as a 3D image, as seen in \cref{fig-app:3dchess} \textit{(right)}, it can be difficult to understand the board state from this visualisation, so we have also produced a standard 2D rendering of the same game in \cref{fig-app:3dchess} \textit{left}.

The dataset has three classes which denote the eventual outcome of the game, corresponding to a white win, draw, and black win. It is for this reason that the very final board states are excluded from the samples to avoid
presenting a state that directly depicts a checkmate and, therefore, the true outcome of the game. A rendered sample of each class can be seen in \cref{App-fig:chess}.

\begin{figure*}
    \centering
  \includesvg[width=\textwidth]{Images/Chesseract.svg}
  
  \caption{Examples of the Chesseract dataset converted into a standard board format.}
  \label{App-fig:chess}
\end{figure*}

\begin{figure*}
    \centering
    \includesvg[width=0.4\textwidth]{Images/Chesseract_3d-2d.svg}\hfill
    \includesvg[width=0.55\textwidth]{Images/Chesseract_3d.svg}
    \caption{As Chesseract uses a 12-channel one-hot encoding to represent a board state, the board can be viewed as a 3D object. \textit{left}: shows an example game that results in a Draw, where we have flipped the board so white is at the bottom as it would be in traditional notation. \textit{right}: shows a 3D rendering of the \textit{left} image. The white pieces are shown as grey in the 3D rendering for easier viewing.}
    \label{fig-app:3dchess}
\end{figure*}

\section{Competition}
\begin{table}[!htb]
    \begin{minipage}{.5\linewidth}
        
        \centering
        \begin{tabular}{l|rr}
            \textbf{Dataset} & \multicolumn{1}{l}{\textbf{Competition}} & \multicolumn{1}{l}{\textbf{Random}} \\ \hline
            Addnist          & 95.06\%                                  & 5\%                                 \\
            Language         & 89.71\%                                  & 10\%                                \\
            MultNIST         & 95.45\%                                  & 10\%                                \\
            CIFARTile        & 73.08\%                                  & 25\%                                \\
            Gutenberg        & 50.85\%                                  & 16.6\%                              \\
            Isabella         & 61.42\%                                  & 25\%                                \\
            GeoClassing      & 96.08\%                                  & 10\%                                \\
            Chesseract       & 62.98\%                                  & 33.3\%                             
        \end{tabular}
        \caption{The best reported results on each of our datasets from competition submissions.}
        \label{tab:best-comp}
    \end{minipage}%
    \begin{minipage}{.5\linewidth}
      \centering
        
        \begin{tabular}{l|lll}
       & \textbf{GeoClassing}      & \textbf{Isabella}         & \textbf{Chesseract}     \\ \hline
    Rank 1 & 92.37\%          & 59.97\%          & 58.20\%        \\
    Rank 2 & 89.62\%          & \textbf{60.14\%} & 58.10\%        \\
    Rank 3 & \textbf{93.34\%} & 46.87\%          & 59.31\%        \\
    Rank 4 & 90.69\%          & 44.93\%          & 58.74\%        \\
    Rank 5 & 87.81\%          & 43.92\%            & \textbf{63.05\%}
    \end{tabular}
    \caption{The raw scores of the top five submissions of the 2023 competition across the three (at the time) unseen datasets.}
    \label{tab:2023-comp}
    \end{minipage} 
\end{table}

These datasets were initially created and used for the ``Unseen Data Challenge" held at the CVPR 2021, 2022, and 2023 NAS workshops.
%The datasets presented in this paper were originally created for challenges run at the Neural Archit. 
The challenges required participants to create NAS techniques that produced models to achieve optimum performance on hidden, unseen datasets.

The participants were given a starting kit to develop their NAS methods and development datasets they could use to aid method construction. Once submitted, these methods were given 24 hours to find optimal models for three datasets, an average of eight hours per dataset.

The competition was scored using the following formula 

$$
\sum_{i=0}^{2}\max(-10, (r_i - b_i) * (10/(100-b_i))),
$$ 

\noindent where $r_i$ is the raw accuracy the model scored on dataset $i$, and $b_i$ is the baseline score (from ResNet-18) on dataset $i$ . We used this formula to encourage generalisable solutions, as using this formula, scores are capped between -30 and 30 (-10 and 10 on individual datasets). This prevents a single dataset from skewing the results too much, either positively or negatively.

\cref{tab:best-comp} contains the highest reported results achieved on each of our datasets from submissions to the competition.

\cref{tab:2023-comp} shows the breakdown of results from the top 5 participants from the competition in 2023, which used GeoClassing, Isabella, and Chesseract as the hidden datasets. It is interesting to note that the overall winner of the 2023 competition's algorithm did not produce any models that gave the top performance. Instead, it proved more generalisable, producing models that, while not excelling on a singular dataset, generate competitive results across all datasets.

The top two ranked submissions to the 2023 competition used an evolutionary CNN NAS approach, adding convolutional blocks throughout the training process. Ranks three, four, and five deployed super-nets trained on Image-Net and applied evolutionary search using random sampling. To preserve the intellectual property of the participants, we cannot reveal the details of their submissions.